# Drone Detection Using Convolutional Neural Networks

Fatemeh Mahdavi
Faculty of Electrical and Computer Engineering
Qom University of Technology (QUT)
Qom, Iran
mahdavi.f@qut.ac.ir

Roozbeh Rajabi
Faculty of Electrical and Computer Engineering
Qom University of Technology (QUT)
Qom, Iran
rajabi@qut.ac.ir

*Abstract*— **In image processing, it is essential to detect and track air targets, especially UAVs. In this paper, we detect the flying drone using a fisheye camera. In the field of diagnosis and classification of objects, there are always many problems that prevent the development of rapid and significant progress in this area. During the previous decades, a couple of advanced classification methods such as convolutional neural networks and support vector machines have been developed. In this study, the drone was detected using three methods of classification of convolutional neural network (CNN), support vector machine (SVM), and nearest neighbor. The outcomes show that CNN, SVM, and nearest neighbor have total accuracy of 95%, 88%, and 80%, respectively. Compared with other classifiers with the same experimental conditions, the accuracy of the convolutional neural network classifier is satisfactory.**

## I. Introduction

With the fast advancement within the field of unmanned vehicles and innovation utilized to build them, the number of drones produced for military, commercial, or recreational purposes increments pointedly as time passes. This circumstance presents urgent protection and security dangers when cameras or weapons are appended to the drones. Thus, detection of the position and properties, similar to speed and track, of drones before an undesirable happening, has gotten to be vital[1].

Speed and move capability of drones, their similarity to birds in appearance when seen from a distance, as shown in Fig. 1 make it challenging to detect, identify, and correctly localize them [2]. because of high variability among the objects of a similar kind, object detection in the physical environment is an onerous duty, furthermore changes in appearance, illumination, and perspective reduce the efficiency of a detector. Most of the detectors perform ineffectively within the case of changes to the scale and distortion[2].

Classifications can be divided into available and advanced categories. General classification methods include Maximum likelihood classification (MLC) and Minimum distance classification (MDC). Advanced classifications, artificial neural networks (ANNs), Decision trees, support vector machines, and object-oriented classifications can also be mentioned [3] .

In general, the classification is done in several stages in such a way that first the required data, which are called train data, are collected in various ways. then the selected classification algorithm is applied to these images, and the whole Images are classified Then the accuracy of the desired classification can be examined using test data [3].

The remains of the paper are as follows: In Section II, it presents a brief of related work. Section III, shows the proposed methods. Analysis of the simulation results is presented in Section IV. Then, some conclusions are discussed in Section V.

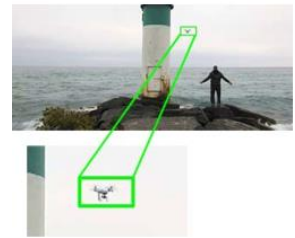

Fig. 1. Drone Detection [1]

## II. RELATED WORK

The author of [4] proposed a drone detection solution based on the Radio Frequency (RF) emitted during the live communication session between the drone and its controller using a Deep Learning (DL) technique, namely, the Convolutional Neural Network (CNN). Mejias et al. used morphological pre-preparing and Hidden Markov Model channels to distinguish and track small unmanned vehicles [5].

In [6], the authors presented a comprehensive review of the literature, addressing the concept of drone detection through object detection and classification using Machine Learning technologies. Essentially, the application of machine learning facilitate the drone detection in a binary classification model as "drone" or "no drone". However, some research in the literature goes beyond the traditional classification to a multiclass classification which identifies drone types.

Goke et al. utilized cascaded boosted classifiers with some local highlight descriptors [7]. Using a Doppler radar system, Michael et al. [8] detected a flying drone in which data would be gathered by a two-channel Doppler radar and used for Range-Doppler-azimuth processing. In [9], the drone is detected using a convolutional neural network by audio signals received by the microphone. Because the drone's harmonic properties are different from objects such as motorcycles, etc. that make similar noises, it has used the Short-Time Fourier Transform (STFT) to extract features. In [10], drone is detected using frequency-modulated continuous wave (FMCW) radar system. The system uses antennas with a frequency band of 11 to 11.15 GHz to produce a linear wave of sawtooth antennas with a gain of 33.7dBi. Metal holes are used between and on both sides of the antenna to minimize path interference. In this method, the drone is detected at a distance of more than 500 meters.

There are several methods for detecting Drones. In this article, machine learning strategies such as, Convolutional Neural Network (CNN), K-Nearest Neighbor (KNN), Support Vector Machine (SVM) are utilized and then compared.

## III. METHODS

As mentioned in this article, machine learning strategies such as, convolutional neural network, support vector machine, and nearest neighbor, have been utilized to detect drones. In this segment, the presentation of the proposed methods will be inspected. The characteristics of the layers used in CNN will also be stated.

### A. K-Nearest Neighbor

The k-nearest neighbor is an optimization issue to discover the nearest dots in metric spaces. The nearest neighbor algorithm is used as the first classification model because this model is simple and is robust against a set of noisy data and is also useful and applicable for a large data set. It is the form that assigns data that does not belong to a category to the group that has the most data around anonymous data.

In Fig. 2, using the 16-NN rule, the point denoted by a black dot is classified into the class of the blue points. Out of the sixteen nearest neighbors, five are red, and eleven are blue. The circle indicates the area within which the sixteen nearest neighbors lie[11].

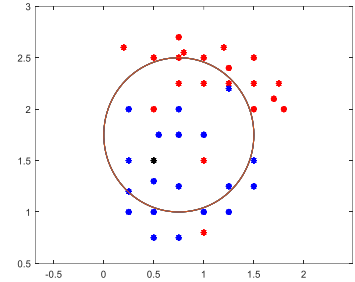

Fig. 2. 16- Nearest Neighbor

### B. Support Vector Machine

In a variety of situations, we need to categorize the information we have. In machine learning, this is called classification. A support vector machine (SVM) is a set of monitoring algorithms that analyzes data. In SVM, some labeled data must be used to classify information [12]. In general, data in two-dimensional space is separated by a line in three-dimensional space by a plane, and in n-dimensional space by a hyperplane.

The Support vector is closest data to the decision boundary, and margin is also the distance of the support vectors to the decision boundary. The goal is maximize the distance of the support vectors from the decision boundary. Relation (1) represent the decision boundary equation that w is slope and b is bias of line. The purpose of determining w and b is not only to classify the data correctly but also to have the decision distance as far as possible with the support vectors which is expressed mathematically in equation (2) and (3) [3]

$$w^t x + b = 0 \quad (1)$$

$$x = (x^t, y^t) \qquad y^t = \begin{cases} +1, & \text{if } x^t \in c_1 \\ -1, & \text{if } x^t \in c_2 \end{cases} \quad (2)$$

$$\begin{cases} w^t x^t + b \geq +1, & \text{for } y_t = +1 \\ w^t x^t + b \geq -1, & \text{for } y_t = -1 \end{cases} \quad (3)$$

X is a point on the decision boundary, and w is vector for determine slope [13] . Thus, by finding w and b, the decision boundary equation can be obtained.

## C. Convolutional Neural Network

Convolutional neural networks (CNN) are used for analysis in machine learning. This method is very efficient and is one of the most common ways in various applications of computer vision.

Neural network architecture is presented in different forms depending on the type and number of layers, that the most common of which are multilayer perceptron [3]. The function of this method is based on Equation (4), in which θ represents the threshold, $w^t$ represents the weight vector of the coefficients, and x represents the input vector [3].

$$g=f(w^t x + \theta) \quad (4)$$

A multilayer perceptron consists of an input layer, a hidden layer, and an output layer, each containing a certain number of neurons. The choice of the number of hidden layers and neurons in each of them is significant because if their number is small, the network will face a lack of learning resources to solve complex problems. if it is high, it will cause Network training time increases [14]. In total, a CNN network comprises of three primary layers: the convolution layer, the pooling layer, and the fully connected layer.

The pooling layer reduces the size, also decreases the number of parameters, and reduces the volume of calculations, which in turn helps to control the fitting phenomenon. In the j map, the $a_{ij}$ attribute can be defined in the ith layer as follows:

$$a_i = f(w_n * a_{i-1} + b_{ij}) \quad (5)$$

Where f (.) is an activation function described as f (x) = max (0, x), it is called the ReLU layer. w is the core of the filters, n is the index of the feature map set that Attached to the feature map in (i-1) layer. ∗ is the convolution operator, and $b_{ij}$ is the bias value of the current feature map. w and b must be trained to be able to extract better features.

In each convolutional neural network, there are two steps to preparing. Feedforward stage and backpropagation stage. In the Feedforward stage, the input image is fed to the system. This operation is nothing but multiplying the point between the input and the parameters of each neuron and finally performing convolutional operations in each layer. The network output is then calculated.

### 1) Proposed CNN Structure

We propose CNN structure as indicated by the multifaceted nature of our information. According to Fig. 3, the system consist of three max-pooling layers, three convolution layers and two fully connected layers. The function applied to the data is soft max. Epoch was also considered 80. The epochs are repeated until the error reaches a fixed value and does not change.

## IV. RESULTS

### A. Datasets

The proposed methods use films in which there are drones and birds. These films are converted into frames by MATLAB software. After extracting the video frames, the ones that do not have a drone and a bird are not used, and in the other frames, we separate the images of the drone and the birds and put them in two separate folders, one folder containing the images of drones and the other include images of birds. As a result of this operation, 712 images of drones and birds were collected. When choosing a set of images, it should be noted that the images in question should not have marginal and unimportant goals as much as possible, because these goals can destroy the performance of the program[15].

Eventually, about 80% of this data is utilized for training and 20% for testing. In this way, images of 500 birds and drones are randomly selected and provided to the algorithm for training at each execution. The residue of the images are used for testing.

This algorithm is implemented in the MATLAB software environment.

TABLE I. HOG FEATURE EXTRACTION FOR SAMPLE DRONE AND BIRD

| Image | Feature Extraction |
|---|---|
| 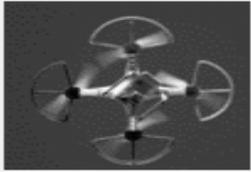 | 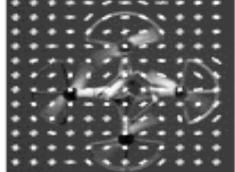 |
| 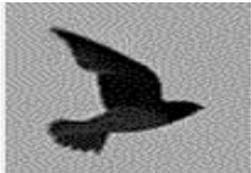 | 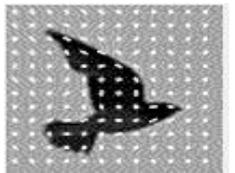 |

### B. Pre processing

Due to the different sizes of the images in the bird and drone data, before executing the main parts of the algorithm, all the images are resized to the same dimensions (48 × 48) to process the algorithm better.

#### 1) Extract feature

The purpose of feature extraction is to transform the image into a set of feature vectors that can describe the image well and be used in the classification and recognition phase.

Table I shows how to extract features from drones and birds in the database for SVM and KNN methods (the directional gradient histogram feature is used [16]). CNN can leave out complex data pre-processing and extract feature automatically [17].

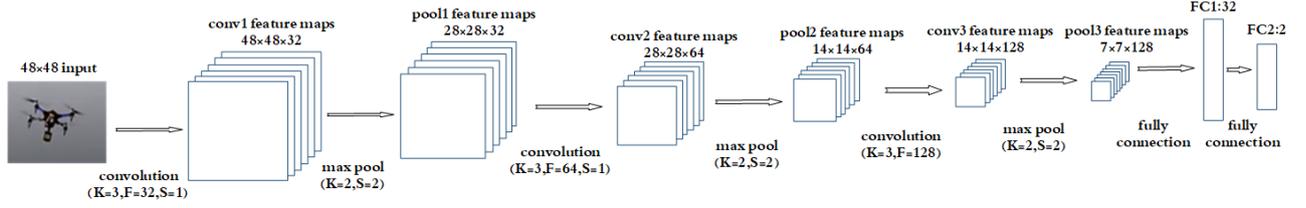

Fig. 3. The proposed CNN structure

Finally, the general process used in this paper is shown in Fig. 4

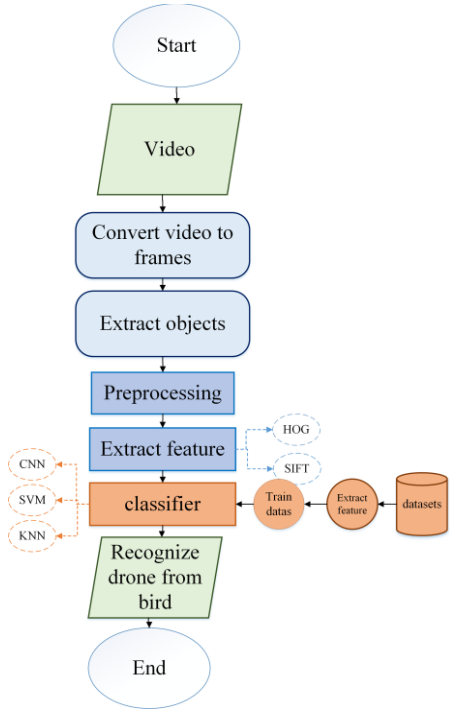

Fig. 4. Flowchart of the proposed drone detection algorithm

## C. Operation of Classification

According to Fig. 5.a, in exceptional cases such as the drone being halfway, since it is very similar to a bird, even the CNN algorithm is not able to detect the drone and misidentifies the bird. Fig. 5.b, because the two birds are very close to each other, the KNN classifier misdiagnoses the drone but is correctly identified by CNN and SVM. As shown in Table II, KNN performs worse than SVM and CNN.

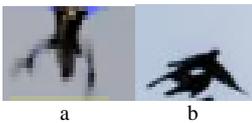

a      b
Fig. 5. Examples of undetectable samples

To test the results, different criteria have been used. These criteria include:

- Accuracy: The number of samples that the system has successfully detected.
- Sensitivity: How many of the available correct samples have been selected.
- Precision: How many of the selected samples are valid. These evaluation metric are given in formulas 6 - 8.

TP :True positive - accurate positive forecast.

TN :True negative - inaccurate positive forecast.

FP :False positive - accurate negative forecast.

FN :False negative - inaccurate negative forecast.

$$\text{Accuracy} = \frac{TP+TN}{TP+FP+FN+TN} \quad (6)$$

$$\text{Sensitivity} = \frac{TP}{TP+FN} \quad (7)$$

$$\text{Precision} = \frac{TP}{TP+FP} \quad (8)$$

Table II shows the classification test using the test data. As it turns out, convolutional neural network classifiers perform better in drone detection than other classifiers.

TABLE II. COMPARISON RESULTS OF THE CLASSIFIERS

| Classifier | Sensitivity (%) | Precision (%) | Accuracy (%) |
|---|---|---|---|
| CNN | ٩3 | ٩7 | ٩5 |
| SVM | ٩١ | ٨٦ | ٨٨ |
| KNN | ٩٤ | ٧٤ | ٨٠ |

Table III shows that the more convolutional layers and Epoch, the more accurate the diagnosis, but the more time is needed for training. It should be noted that this increase will improve performance to some extent, will not have much effect on program performance.

TABLE III. COMPARISON RESULTS OF CNN BASED ON THE NUMBER OF LAYERS

| CNN Properties | two Conv layer | four Conv layer |
|---|---|---|
| | 60 Epoch | 80 Epoch |
| Time | 9min, 56 sec | 14 min, 28sec |
| Accuracy (%) | 85 | 95 |

The specifications of the system with which the above results are obtained are as follows:

- Intel Core i5-4200M (2.5GHZ)
- 2 GB DDR3 L Memory
- 1TB HDD

For example, Fig. 6 shows a video frame in which the drone is correctly separated from the birds.

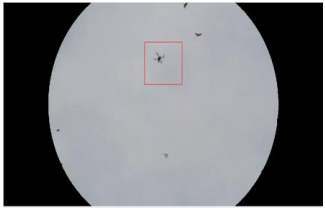

Fig. 6. Drone Detection Using CNN

## V. CONCLUSION

Today, drones are used for various purposes. These include goals such as protection, passage to small areas, espionage uses, and so on. The small size and low noise of the drone make its role more critical. Therefore, drone detection has also become important for security issues. In this paper, the proposed algorithm has two stages of training and testing. In the training phase, the properties of all the images used were extracted, and a convolutional neural network was trained using these properties. Then, in the next step of experiment, the input image was classified by a trained system. Therefore, the confusion between some similar parts of the drone and the bird was eliminated using the convolutional neural network classification model. The convolutional neural network model improves the efficiency in comparison of the support vector machine and the nearest neighbor in detecting similar parts of the two objects.